\documentclass[10pt,twocolumn,letterpaper]{article}

\usepackage{cvpr}
\usepackage{times}
\usepackage{epsfig}
\usepackage{graphicx}
\usepackage{amsmath}
\usepackage{amssymb}
\usepackage{booktabs}
\usepackage{capt-of,etoolbox}

\usepackage[pagebackref=true,breaklinks=true,letterpaper=true,colorlinks,bookmarks=false]{hyperref}

\cvprfinalcopy

\def\cvprPaperID{***}

\newcommand{\R}{\mathbb{R}}
\newcommand{\T}{\mathsf{T}}
\renewcommand{\paragraph}[1]{\vspace{1mm}\noindent\textbf{#1}}
\begin{document}

\title{Shape from Shading through Shape Evolution}

\author{Dawei Yang \quad Jia Deng \\
Computer Science and Engineering, University of Michigan \\
\texttt{\small\{ydawei,jiadeng\}@umich.edu} \\
}

\maketitle

\begin{figure*}
    \centering
    \includegraphics[width=\linewidth]{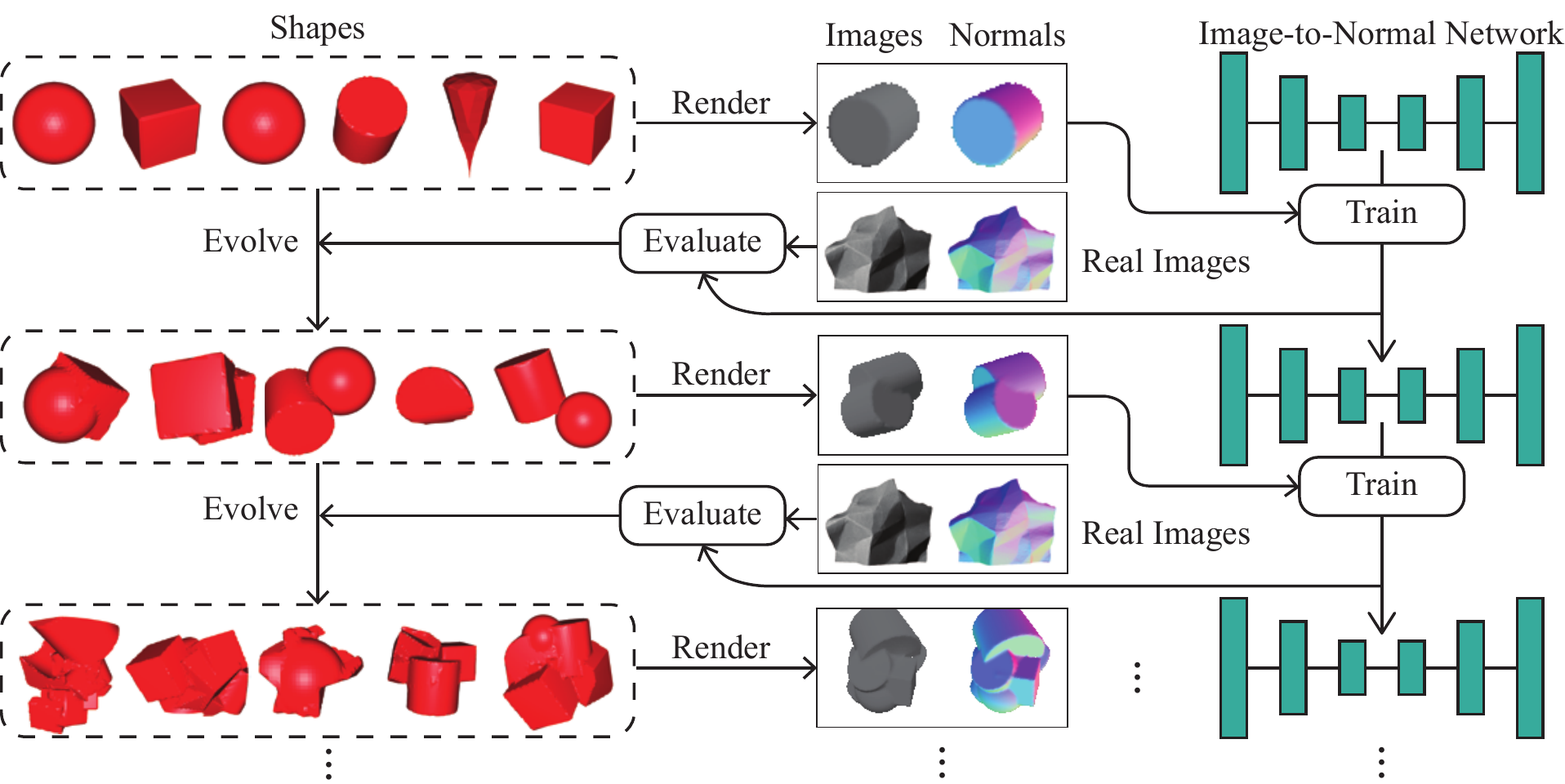}
    \caption{The overview of our approach. Starting from simple primitives such as spheres and cubes, we evolve a population of complex shapes.
    We render synthetic images from the shapes to incrementally train a shape-from-shading network.
    The performance of the network on a validation set of real images is then used to guide the shape evolution.
    }
    \label{fig:teaser}
\end{figure*}

\begin{abstract}

    In this paper, we address the shape-from-shading problem by training deep networks
    with synthetic images. Unlike conventional approaches that combine deep learning and
    synthetic imagery, we propose an approach that does not need any external shape
    dataset to render synthetic images. Our approach consists of two synergistic 
    processes: the evolution of complex shapes from simple primitives, and the training of
    a deep network for shape-from-shading. The evolution generates better shapes guided by
    the network training, while the training improves by using the evolved shapes. We show
    that our approach achieves state-of-the-art performance on a shape-from-shading
    benchmark. 
    
\end{abstract}

\section{Introduction}

Shape from Shading (SFS) is a classic computer vision problem 
at the core of single-image 3D reconstruction ~\cite{zhang1999shape}.
Shading cues play an important role in recovering geometry and are
especially critical for textureless surfaces.

Traditionally, Shape from Shading has been approached as an
optimization problem where the task is to solve for a plausible shape
that can generate the pixels under a Lambertian shading model~\cite{xiong2015shading,ecker2010polynomial,BarronTPAMI2015,barron2012shape,barron2011high}. The key
challenge is to design an appropriate optimization objective to sufficiently constrain
the solution space, and to design an optimization algorithm to
find a good solution efficiently.

In this paper, we address Shape from Shading by training deep networks
on synthetic images. This follows an emerging line of work on single-image 3D
reconstruction that
combines synthetic imagery and deep learning~\cite{su2015render,mccormac2017scenenet,massa2016deep,zhang2016physically,richter2017playing,tatarchenko2016multi,choy20163d,xiang2016objectnet3d,butler2012a}.
Such an approach does away with the manual design of optimization objectives and algorithms, and instead
trains a deep network to directly estimate shape. This approach can take advantage of a large
amount of training data, and has shown great
promise on tasks such as view point estimation~\cite{su2015render},
3D object reconstruction and recognition~\cite{tatarchenko2016multi,choy20163d,xiang2016objectnet3d},
and normal estimation in indoor scenes~\cite{zhang2016physically}.

One limitation of this data-driven approach, however, is availability of
3D shapes needed for rendering synthetic images. Existing approaches have relied on
 manually constructed~\cite{shapenet2015v1,wu20153d,aubry2014seeing} or scanned shapes~\cite{choi2016a}.
But such datasets can be expensive to build. Furthermore, while synthetic datasets can be
augmented with varying viewpoints and lighting, they are still constrained by the number
of distinct shapes, which may limit the ability of trained models to generalize to real
images.

An intriguing question is whether it would be possible to do away with manually curated 3D shapes
while still being able to use synthetic images to train deep networks. Our key
hypothesis is that shapes are compositional and we should be able to compose complex
shapes from simple primitives. The challenge is how to enable automatic composition and
how to ensure that the composed shapes are useful for training deep networks.

We propose an evolutionary algorithm that jointly generates 3D shapes and trains 
a shape-from-shading deep network. We evolve complex shapes entirely from simple
primitives such as spheres and cubes, and do so in tandem with the training of a deep
network to perform shape from shading. The evolution of shapes and the training of a deep
network collaborate---the former generates shapes needed by the latter, and the latter
provides feedback to guide the former. 
Our approach is significantly novel compared to prior works that use synthetic images to train deep
networks, because they have all relied on manually curated shape datasets~\cite{su2015render,massa2016deep,zhang2016physically,tatarchenko2016multi}.

In this algorithm, we represent each shape using an implicit function~\cite{ricci1973constructive}.
 Each function is composed of simple primitives, and the composition is encoded as
a computation graph. Starting from simple primitives such as spheres and cubes, we
evolve a population of shapes through transformations and compositions defined over graphs. 
We render synthetic images from each shape in the population and
use the synthetic images to train a shape-from-shading network. The performance of the
network on a validation set of real images is then used to define the fitness score of each shape. In each round of the
evolution, fitter shapes have better chance of survival whereas less fit shapes tend to be
eliminated. The end result is a population of surviving shapes, along with a
shape-from-shading network trained with them. Fig.~\ref{fig:teaser}
illustrates the overall pipeline.

The shape-from-shading network is incrementally trained in a way that is tightly
integrated with shape evolution. In each round of evolution, the network is fine-tuned \emph{separately} with each 
shape in the population, spawning one new network instance per shape. Then the best
network instance advances to the next round while the rest are
discarded. In other words, the network tries updating its weights using each newly evolved shape, and the best
weights are kept to the next round. 

We evaluate our approach using the MIT-Berkeley Intrinsic Images
dataset~\cite{BarronTPAMI2015}. Experiments demonstrate
that we can train a deep network to achieve state-of-the-art performance on real
images using synthetic images rendered entirely from evolved shapes, without the help of 
any manually constructed or scanned shapes. In addition, we present ablation studies which support
the design of our evolutionary algorithm. 

Our results are significant in that we demonstrate that it is potentially possible to
completely automate the generation of synthetic images used to train deep networks.
We also show that the generation procedure can be
effectively adapted, through evolution, to the training of a deep
network. This opens up the possibility of training 3D reconstruction networks with a large
number of shapes beyond the reach of manually curated shape collections. 

To summarize, our contributions are twofold: (1) we propose an evolutionary algorithm
to jointly evolve 3D shapes and train deep networks, which, to the best of our knowledge,
is the first time this has been done; (2) we demonstrate that a network trained this way
 can achieve state-of-the-art performance on a real-world shape-from-shading benchmark,
 without using any external dataset of 3D shapes. 

\section{Related Work}
Recovering 3D properties from a single image is one of the
most fundamental problems of computer vision. Early works mostly focused on developing
analytical solutions and optimization techniques, with zero or minimal
learning~\cite{criminisi2000shape,zhang1999shape,BarronTPAMI2015,barron2012shape,barron2011high}.
Recent successes in this direction include the SIRFS algorithm by Barron and
Malik~\cite{BarronTPAMI2015}, the local shape from shading method by Xiong
\etal~\cite{xiong2015shading}, and ``polynomial SFS'' algorithm by Ecker and
Jepson~\cite{ecker2010polynomial}.  All these methods have interpretable, ``glass box''
models with elegant insights, but in order to maintain analytical tractability, they have
to make substantial assumptions that may not hold in unconstrained settings.  For example,
SIRFS~\cite{BarronTPAMI2015} assumes a known object boundary, which is often unavailable
in practice. The method by Xiong \etal assumes quadratically parameterized surfaces, which
has difficulties approximating sharp edges or depth discontinuities.

Learning-based methods are less interpretable but more
flexible.  Seminal works include an MRF-based method proposed by Hoiem
\etal~\cite{hoiem2007recovering} and the Make3D~\cite{saxena2009make3d} system by Saxena
\etal.  Cole \etal~\cite{cole2012shapecollage} proposed a data-driven method for 3D shape
interpretation by retrieving similar image patches from a training set and stitching the
local shapes together. Richter and Roth~\cite{richter2015discriminative} used a
discriminative learning approach to recover shape from shading in unknown illumination.
Some recent works have used deep neural networks for predicting surface
normals~\cite{wang2015designing,Bansal2016} or
depth~\cite{Eigen2015,wang2015towards,chakrabarti2016depth} and have shown
state-of-the-art results. 

Learning-based methods cannot succeed without high-quality training data. Recent years have
seen many efforts to acquire 3D ground truth from the real world, including 
 ScanNet~\cite{dai2017scannet}, 
 NYU Depth~\cite{Silberman:ECCV12}, the KITTI Vision Benchmark
 Suite~\cite{Geiger2012CVPR}, SUN RGB-D~\cite{song2015sun}, B3DO~\cite{janoch2013category},
 and Make3D~\cite{saxena2009make3d}, all of which offer RGB-D images captured by depth
sensors. The MIT-Berkeley Intrinsic Images dataset~\cite{BarronTPAMI2015}
 provides real world images with ground
 truth on shading, reflectance, normals in addition to depth. 

In addition to real world data, synthetic imagery has also been explored 
 as a source of supervision. Promising results have been
demonstrated on diverse 3D tasks such as pose estimation~\cite{su2015render,massa2016deep,aubry2014seeing}, optical 
flow~\cite{butler2012a}, object reconstruction~\cite{tatarchenko2016multi,choy20163d}, and
surface normal estimation~\cite{zhang2016physically}. Such advances have been made
possible by concomitant efforts to collect 3D content needed for rendering. In particular, 
the 3D shapes have come from a
variety of sources, including online CAD model
repositories~\cite{shapenet2015v1,wu20153d}, interior design
sites~\cite{zhang2016physically}, video
games~\cite{richter2017playing,richter2016playing}, and
movies~\cite{butler2012a}.

The collection of 3D shapes, from either the real world or a virtual world, involves
substantial manual effort---the former requires depth sensors to be carried around whereas
the latter requires human artists to compose the 3D models. Our work explores a
new direction that automatically generates 3D shapes to serve an end task, bypassing real
world acquisition or human creation. 

Our work draws inspiration from the work of Clune \& Lipson~\cite{clune2011evolving},
which evolves 3D shapes as Compositional Pattern Producing
Networks~\cite{stanley2007compositional}. Our work differs from theirs in
two important aspects. First, Clune \& Lipson perform only shape generation, particularly
the generation of interesting shapes, where interestingness is defined by humans in the
loop. In contrast, we
\emph{jointly} generate shapes and train deep networks, which, to the best of our knowledge, is the first
this has been done. Second, we use a significantly different evolution procedure. Clune \&
Lipson adopt the NEAT algorithm~\cite{stanley2002evolving}, which uses generic graph operations
such as insertion and crossover at random nodes, whereas our evolution
operations represent common shape ``edits'' such as translation, rotation, 
intersection, and union, which are chosen to optimize the efficiency of evolving 3D
shapes.

\section{Shape Evolution}

Our shape evolution follows the setup of a standard genetic
algorithm~\cite{holland1975adaptation}. We start with an initial population of shapes. Each shape in the
population receives a fitness score from an external evaluator. Then the shapes are
sampled according to their fitness scores, and undergo random geometric operations
to form a new population. This process then repeats for successive iterations. 

\subsection{Shape Representation}

We represent shapes using implicit surfaces~\cite{ricci1973constructive}. An implicit surface is defined by
a function $F: \R^3 \rightarrow \R.$ that maps a 3D point to a scalar.
The surface consists of points $(x,y,z)$ that satisfy the equation: 
\[
F(x, y, z) = 0.
\]
And if we define the points $F(x, y, z) < 0$ as the interior,
then a solid shape is constructed from this function $F$.
Note that the shape is not guaranteed to be closed, i.e., may have points at infinity. A simple workaround is to always
confine the points within a cube~\cite{clune2011evolving}.

Our initial shape population consists of four common shapes---
sphere, cylinder, cube, and cone, which can be represented by the functions below: 
\begin{equation}\label{eq:primitive}
\begin{array}{ll}
   \text{Sphere}: & F(x, y, z) = x^2 + y^2 + z^2 - R^2 \\
   \text{Cylinder}: &F(x, y, z) = \max\left(\frac{x^2+y^2}{R^2}, \frac{|z|}{H}\right) - 1 \\
   \text{Cube}: &F(x, y, z) = \max(|x|, |y|, |z|) - \frac L2 \\
   \text{Cone}: &F(x, y, z) = \max(\frac{x^2+y^2}{R^2}-\frac{z^2}{H^2}, -z, z-H)
\end{array}
\end{equation}

An advantage of implicit surfaces is that the composition of shapes can be easily
expressed as the composition of functions, and a composite function can be represented by
a (directed acyclic) computation graph, in the same way a neural network is represented as
a computation graph. 

Suppose a computation graph $G=(V,E)$. It includes a set of nodes 
$V = \{x, y, z\} \cup \{v_1, v_2, \cdots\} \cup \{t\}$, which includes
three input nodes $\{x,y,z\}$, a variable number of internal nodes $\{v_1,v_2, \cdots\}$, and a
single output node $t$. Each node $v\in V$ (excluding input nodes) is associated with a
scalar bias $b_v$, a reduction function $r_v$ that maps a variable number of real values
to a single scalar, and an activation function $\phi_v$ that maps a real value to a
new value. In addition, each edge $e\in E$ is associated with a weight $w_e$. 

It is worth noting that different from a standard neural network or a Compositional
Pattern Producing Network (CPPN) that only uses
\texttt{sum} as the reduction function, our reduction function can be \texttt{sum}, \texttt{max} or
\texttt{min}. As will become clear, this is to allow straightforward composition of
shapes.

\begin{figure}[t]
\centering
\includegraphics[width=\columnwidth]{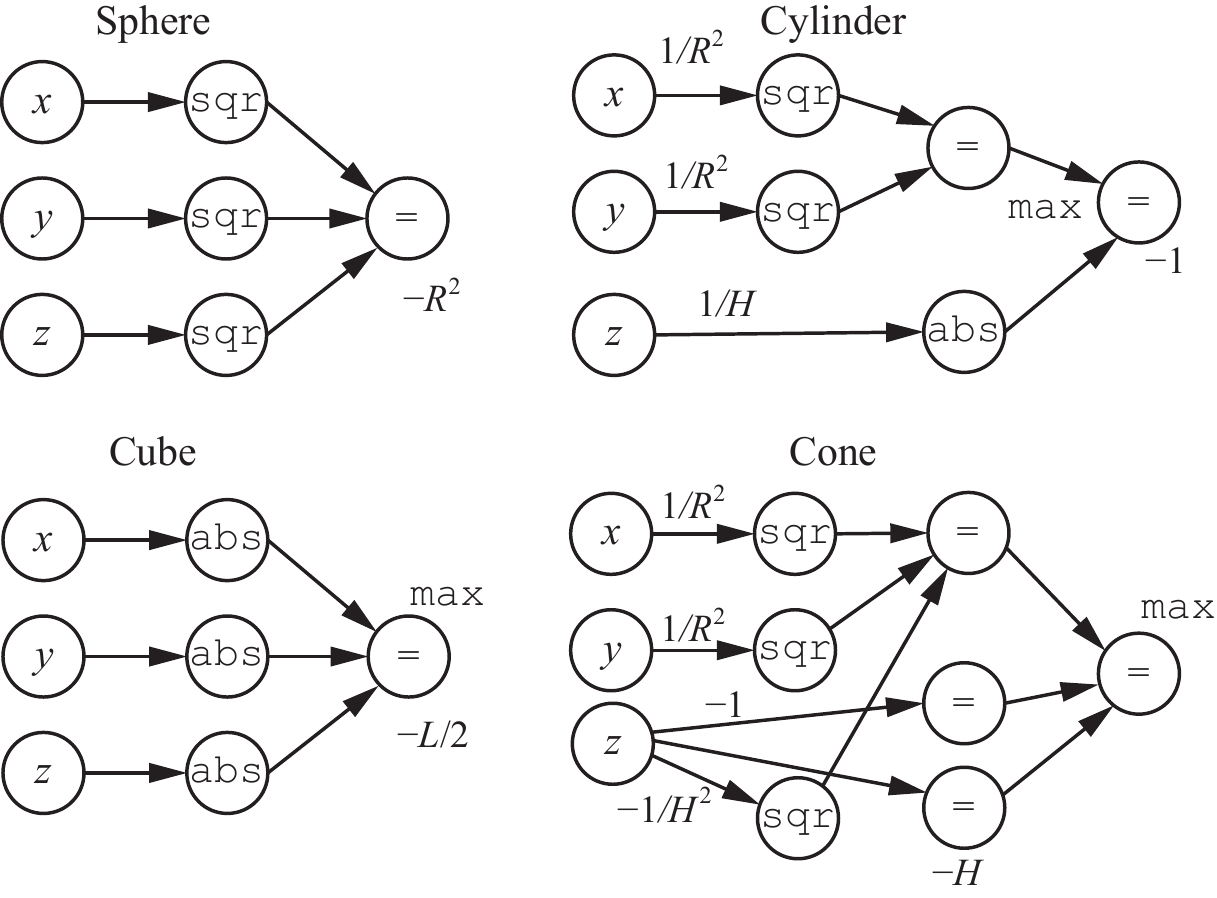}
\caption{The computation graphs of four primitive shapes defined in Eq.~\ref{eq:primitive}.
The unlabeled edge weight and node bias are $0$, and the unlabeled reduction function is \texttt{sum}.}
\label{fig:primitive_shapes}
\end{figure}

To evaluate the computation graph, each node takes the weighted activations of its
predecessors and applies the reduction function, followed by the activation function plus
the bias. 
Fig.~\ref{fig:primitive_shapes} illustrates the graphs of the functions defined in 
Eq.~\ref{eq:primitive}. 

\paragraph{Shape Transformation}
\begin{figure}
\centering
\includegraphics[width=\columnwidth]{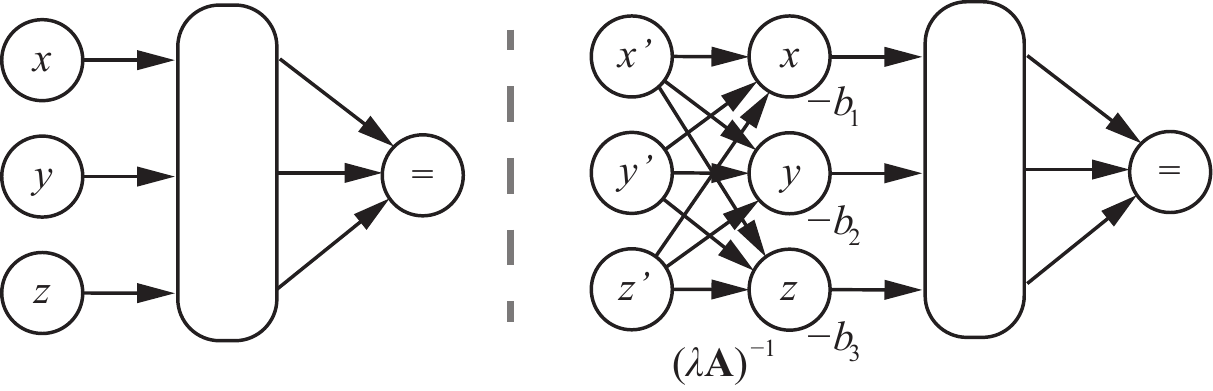}
\caption{Shape transformation represented by graph operation.
Left: the graph of the shape before transformation.
Right: the graph of the shape after transformation.}
\label{fig:shape_transformation}
\end{figure}
To evolve shapes, we define graph operations to generate new shapes from existing
ones. We first show how to transform an individual shape. Given an existing shape
represented by $F(x,y,z)$, let $F(T(x,y,z))$ represent a transformed shape, where 
$T: \mathbb{R}^3\rightarrow
\mathbb{R}^3$ is a 3D-to-3D map. It is easy to verify that $F(T(x,y,z))$ represents the
transformed shape under translation, rotation, and scaling if we define $T$
as
$$
T(x, y, z) = (\lambda\mathbf{A})^{-1}[x,y,z]^\T - \mathbf{b} \text{,}
$$
where $\mathbf{A}$ is a rotation matrix, $\lambda$ is the scalar, and the
$\mathbf{b}$ is the translation vector. Note that for simplicity our definitions have only included a
single global scalar, but more flexibility can be easily introduced by allowing different
scalars along different axes or an arbitrary invertible matrix $\mathbf{A}$. 

This shape transformation can also be expressed in terms of a graph transformation, as
illustrated in Fig.~\ref{fig:shape_transformation}. Given
the original graph of the shape, we insert 3 new input nodes $x',y',z'$ before the
original input nodes, connect new nodes to the original nodes with
weights corresponding to the elements of the matrix $(\lambda\mathbf{A})^{-1}$, and 
set the biases of the original nodes to the vector $-\mathbf{b}$. 

\paragraph{Shape Composition}
\begin{figure}[t]
\centering
\includegraphics[width=\columnwidth]{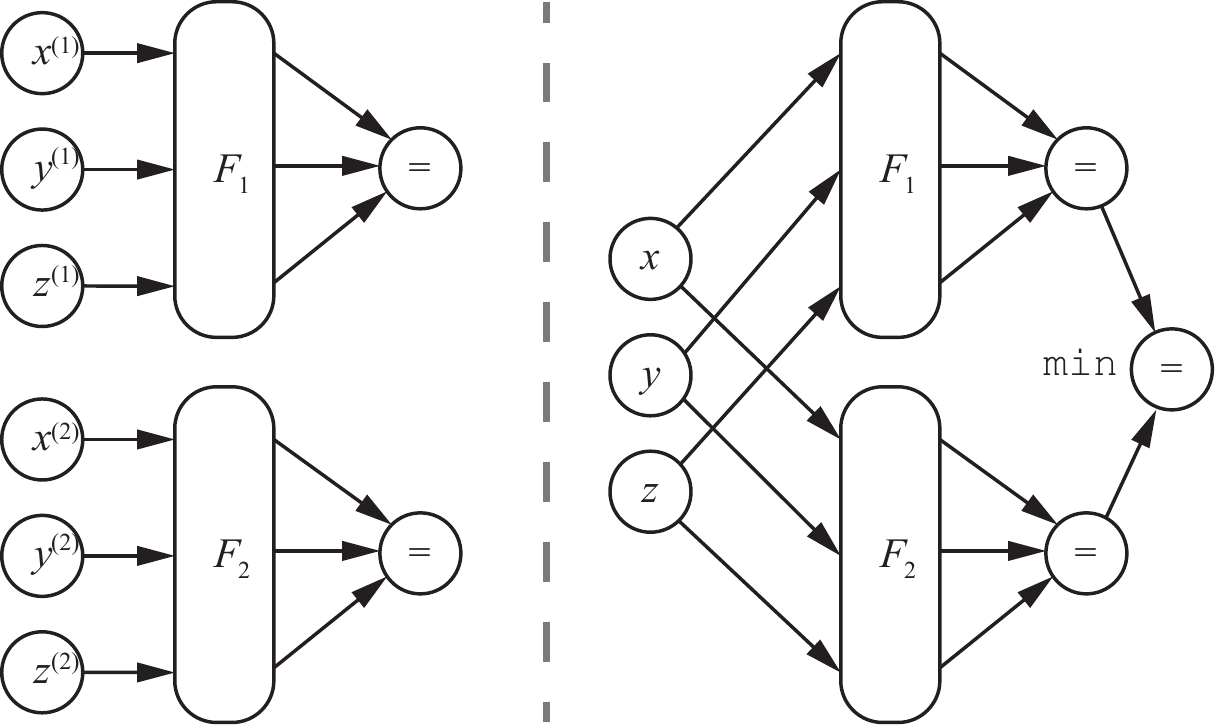}
\caption{The union of two shapes represented by graph merging.
  Left: the respective graphs of the two shapes to be unioned.
  Right: the graph of the unioned shape.
}
\label{fig:shape_composition}
\end{figure}

In addition to transforming individual shapes, we also define binary operations over two shapes.
 This allows complex shapes to emerge from the composition of simple ones. 
Suppose we have two shapes with the implicit representations $F_1(x, y, z)$ and $F_2(x, y, z)$.
As a basic fact~\cite{ricci1973constructive}, the union, intersection, and difference of
the two can be represented as follows:
\begin{align*}
\begin{split}
F_\mathrm{union}(x, y, z) &= \min(F_1(x, y, z), F_2(x, y, z)) \\
F_\mathrm{intersection}(x, y, z) &= \max(F_1(x, y, z), F_2(x, y, z)) \\
F_\mathrm{difference(1,2)}(x, y, z) &= \max(F_1(x, y, z), -F_2(x, y, z)) \text{.}
\end{split}
\end{align*}

In terms of graph operations, composing two shapes together corresponds to merging two graphs. As
illustrated by Fig.~\ref{fig:shape_composition}, we merge the input nodes of the two
graphs and add a new output node that is connected to the two original output nodes. We
set the reduction function (\texttt{max}, \texttt{min}, or \texttt{sum}) and the weights
of the incoming edges to the new output node according to the specific composition
chosen. 

\subsection{Evolution Algorithm} \label{sec:evolution_algorithm}
Our evolution process follows a standard setup. 
It starts with an initial population of $n$ shapes: $\{s_1,s_2,\cdots, s_n\}$, all of
which are primitive shapes described in Eq.~\ref{eq:primitive}.  Next,  $m$
new shapes ($\{s'_1,s'_2,\cdots,s'_m\}$) are created from two randomly sampled existing
shapes (i.e., two parent shapes). Specifically, the two parent shapes each undergo a random
rotation, a random scaling and a random translation, and are then combined by a
random set operation chosen from union, intersection and difference to generate a new
child shape.  Now, the population consists of a total of $n+m$ ($n$ parent shapes and $m$
child shapes). Each shape is then evaluated and given a fitness score, based on which
$n$ shapes are selected  to form the next population. This process is then repeated to evolve more complex shapes. 

Having outlined
the overall algorithm, we now discuss several specific designs we introduce to make our
evolution more efficient and effective. 

\paragraph{Fitness propagation}
Simply evaluating fitness as a function of individual shape is suboptimal in our case.  Our
shapes are evolved based on composition, and to generate a new shape requires combining
existing shapes. If we define fitness strictly on an individual basis, simple shape
primitives, which may be useful in producing more complex shapes, can be eliminated during
the early rounds of evolution. For example, suppose our goal is to evolve an implicit
representation of a target shape.  As the population nears the target shape, smaller and
simpler cuts and additions are needed to further refine the population.  However, if
small, simple shapes, which poorly represent the target shape, have been eliminated, such
refinement cannot take place.

We introduce fitness propagation to combat this problem. We propagate fitness scores from
a child shape to its parents to account for the fact 
that a parent shape may not have a high fitness in itself, but nonetheless should remain
in the population because it can be combined with others to yield good shapes. 
Suppose in one round of evolution, we evaluate each of the $n$ existing shapes and $m$ newly composed
shapes and obtain $n+m$ fitness scores $\{f_1, \cdots, f_n, f'_1, \cdots, f'_m\}$.  But
instead of directly assigning the scores, we propagate the $m$ fitness scores of the child shapes back to the
parent shapes.  A parent shape $f_i$ is assigned the best fitness score obtained by its
children and itself:
$$
f_i \leftarrow \max\left(\{f'_j: s_i \in \pi(s'_j)\}\cup
\{f_i\}\right), 
$$
where $\pi(s'_j)$ is the parents of shape $s'_j$.

\paragraph{Computational resource constraint}

Because shapes evolve through composition, in the course of evolution the shapes will
naturally become more complex and have larger computation graphs. It is easy to
verify that the size of the
computational graph of a composed shape will at least double in the subsequent population.
Thus without any constraint, the average computational cost of a shape will grow
exponentially in the number of iterations as the population evolves, quickly depleting available
computing resources before useful shapes emerge. To overcome this issue, we impose a 
resource constraint by capping the growth of the graphs to be linear in the number of
rounds of evolution. If the number of nodes of a computation graph 
exceeds $\beta t$, where $\beta$ is a hyperparameter and $t$ is time, the graph will be
removed from the population and will not be used to 
construct the next generation of shapes.

\paragraph{Discarding trivial compositions}
A random composition of two shapes can often result in trivial combinations. 
For instance, the intersection of shape $A$ and shape $B$ may be empty, and 
the union of two shapes can be the same as one of the parent. 
We detect and eliminate such cases to prevent them from slowing down the evolution. 

\paragraph{Promoting Diversity}
Diversity of the population is important because it prevents the evolution process from
overcommitting to a narrow range of directions. If the externally given fitness
score is the only criterion for selection, shapes deemed less fit at the moment tend to go extinct 
quickly, and evolution can get stuck due to a homogenized
population. Therefore, we incorporate a diversity constraint into our 
algorithm: a fixed proportion of the shapes in the population are sampled not based on fitness, 
but based on the size of their computation graph, with bigger shapes sampled
proportionally less often.

\section{Joint Training of Deep Network}

The shapes are evolved in conjunction with training a deep network to perform
shape-from-shading.  The network takes a rendered image as input, and predicts
the surface normal at each pixel. To train this network, we render synthetic images and
obtain the ground truth normals using the evolved shapes. 

The network is trained incrementally with a training set that consists of evolved
shapes. Let $D_{i}$ be the training set after the $i$th iteration of the evolution, and
let $N_{i}$ be the network at the same time.  
The training set is initialized to empty before the evolution starts, i.e.\@ $D_0
=\emptyset$, and the network is
initialized with random weights. 

In the $i$th
evolution iteration, 
to compute the fitness
score of a shape $d$ in the population, we \emph{fine-tune} the current network $N_{i-1}$
with $D_{i-1} \cup \{d\}$---the current training set plus the shape in
consideration---to produce a fine-tuned network $N^d_{i-1}$, which is evaluated 
on a validation set of real images to produce an error metric that is then used to define
the fitness score of shape $d$. After we have evaluate the fitness of every shape in the
population, we update the training set with the fittest shape 
 $d^*_i$, 
\[
D_{i} = D_{i-1} \cup \{d^*_i\},
\]
and set the $N^{d^*}_{i-1}$ as the current network, 
\[
N_{i} = N^{d^*}_{i-1}.
\]

In other words, we maintain a growing training set for the network. In each evolution
iteration, for each shape in the population we evaluate what would happen if add the shape
to the training set and continue to train the network with the new training set. This is
done for each shape in the population separately, resulting in as many new network
instances as there are shapes in the current population. The best shape is then officially
added to the training set, and the corresponding fine-tuned network is also kept while the
other network instances are discarded.

\section{Experiments}

\subsection{Standalone Evolution}
We first experiment with shape evolution as a standalone module and study the role of
several design choices. Similar to ~\cite{clune2011evolving}, we evaluate whether our
evolution process is capable of generating shapes close to a given target shape. We define
the fitness score of an evolved shape as its intersection over union (IoU) of volume with
the target shape. 

\paragraph{Implementation details} To select the shapes during evolution, half of the population are sampled
based on the rank $r$ of their fitness score (from high to low), with the selection
probability set to
$0.2^r$. The other half of the
population are sampled based on the rank $s$ of their computation graph size (from small to
large), with the relative selection probability set to $0.2^s$, in order to maintain diversity. 
To compute the volume, we voxelize the shapes to
$32\times32\times32$ grids.
The population size $n$ is $1000$ and the number of child shapes $m=1000$.

\begin{figure}[t]
\centering
\includegraphics[width=\columnwidth]{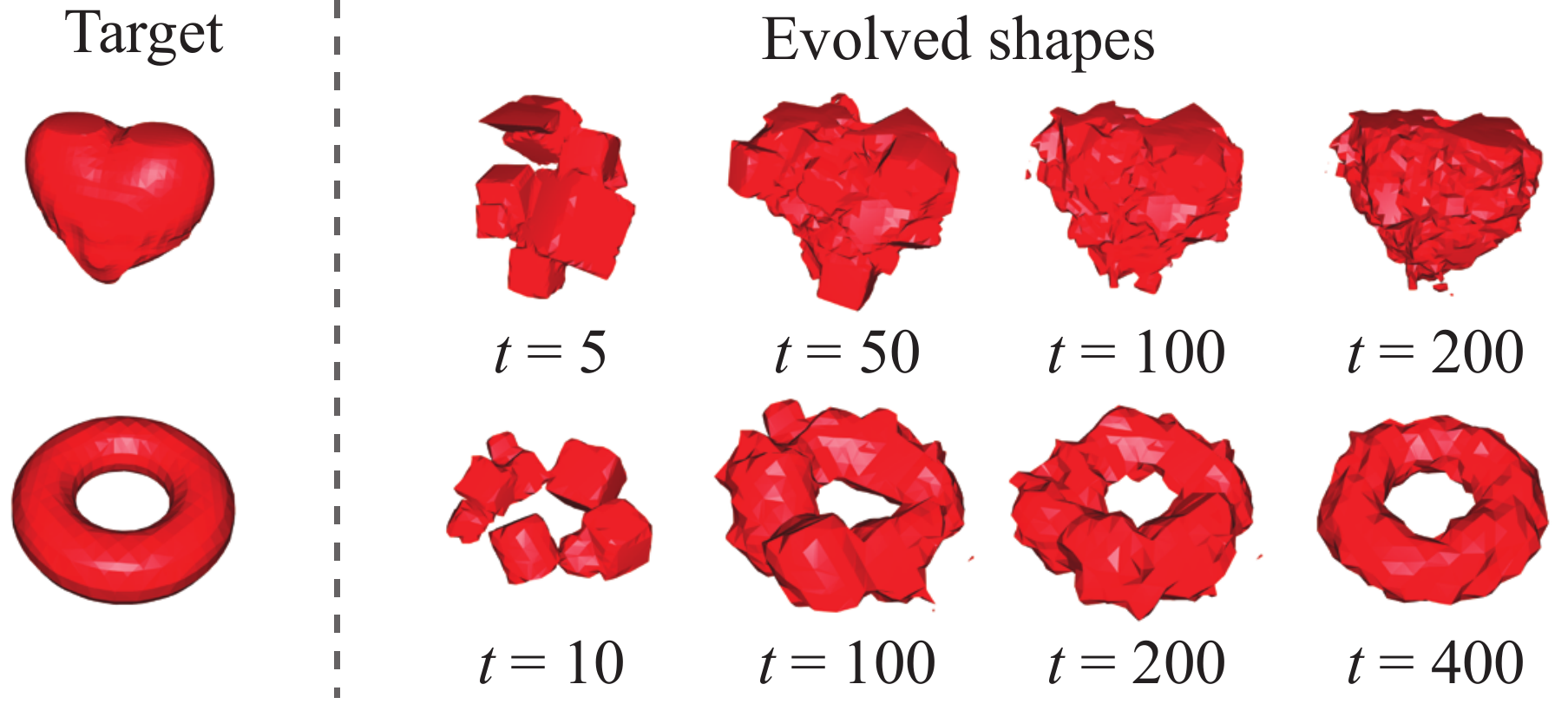}
\caption{Evolution towards a target shape.
Left: targets. Right: The fittest shapes in the population as
the evolution progresses at different iterations.
}
\label{fig:target_evolution}
\end{figure}

\paragraph{Results} 
We use two target shapes, a heart and a torus. Fig.~\ref{fig:target_evolution} shows the two target shapes
along with the fittest shape in the population as the evolution progresses. We can see
that the evolution is able to produce shapes very close to the targets. Quantitatively,
after around 600 iterations, the best IoU of the evolved shapes reaches $94.9\%$ for the
heart  and $93.5\%$ for the torus. 

\begin{figure}[t]
    \centering
    \includegraphics[width=\columnwidth]{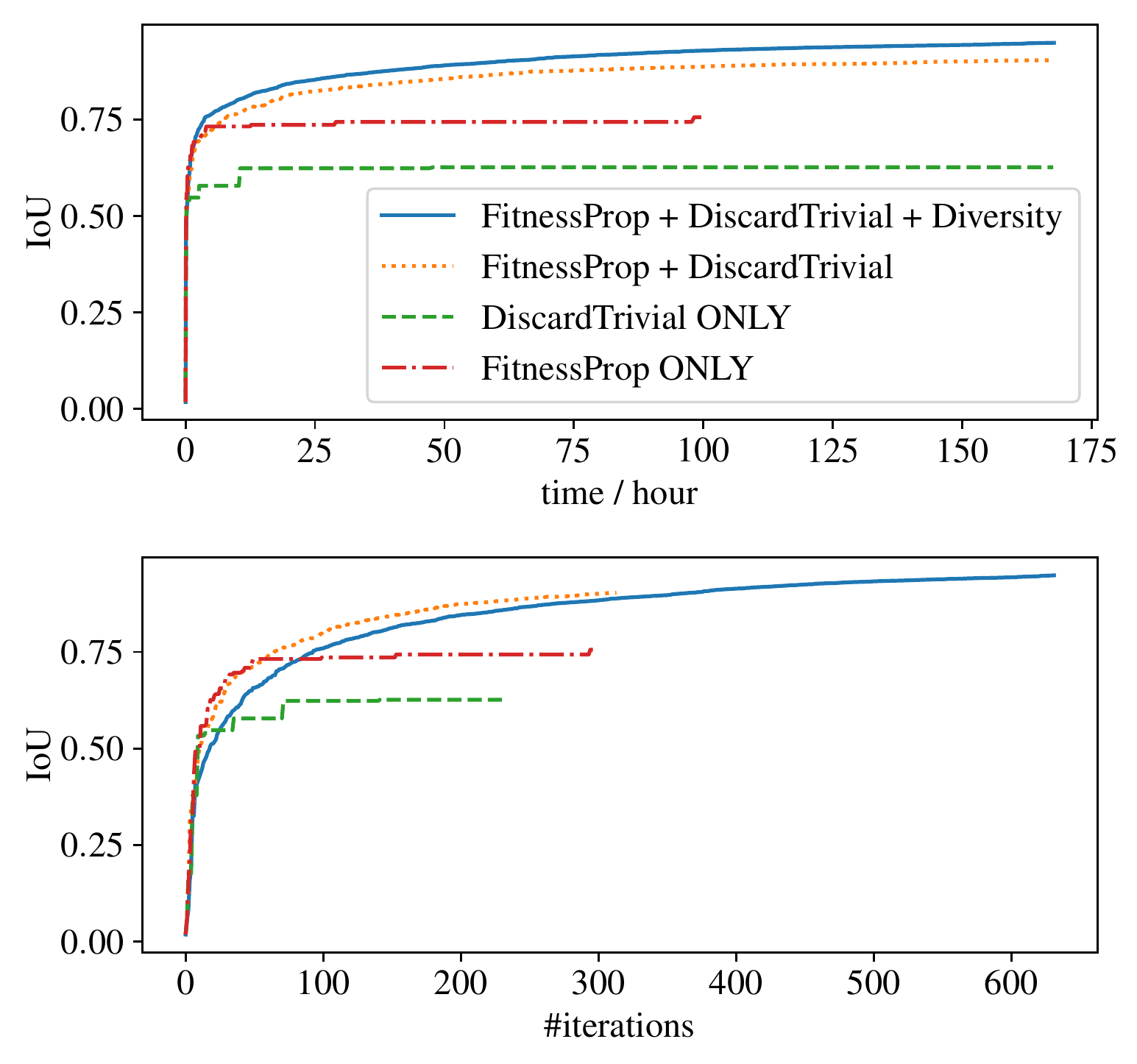}
    \caption{The best IoU with the target shape (heart) versus evolution time (top) and the number of iterations (bottom)
      for different combinations of design choices.
    }
    \label{fig:target_evolution_curve}
\end{figure}

We also study the effect of the design choices described in Sec.~\ref{sec:evolution_algorithm},
including fitness propagation, discarding trivial compositions, and promoting diversity. 
Fig.~\ref{fig:target_evolution_curve} plots, for different combinations of these choices, 
 the best IoU with the target shape (heart) versus
evolution time, in terms of both wall time and the number of iterations. We can see
that each of them is beneficial and enabling all three achieves fastest evolution in terms
of wall time. Note that,
the diversity constraint slows down evolution initially in terms of 
the number of iterations, but it prevents early saturation and is faster
in terms of wall time because of lower computational cost in each iteration. 

\subsection{Joint Evolution and Training}

We now evaluate our full algorithm that jointly evolves shapes and trains a deep
network. We first describe in detail the setup of our individual components. 

\paragraph{Setup of network training}
We use a stacked hourglass
network~\cite{newell2016stacked} as our shape-from-shading network. 
The network consists of a stack of 4 hourglasses, with 16 feature channels for each hourglass
and 32 feature channels for the initial layers before the hourglasses. In each round of
 evolution, we fine-tune the network for $\tau=100$ iterations using
RMSprop~\cite{tieleman2012lecture}, a batch size of 4, and the mean angle error as the
loss function. Before fine-tuning on the new dataset, we re-initialize the RMSprop optimizer.

\paragraph{Rendering synthetic images} To render shapes into synthetic images, we use the
Mitsuba renderer~\cite{Mitsuba}, a physically based photorealistic renderer.
We run the marching cubes algorithm~\cite{lorensen1987marching} on the implicit function of a shape with a resolution of
$64\times64\times64$ to generate the triangle mesh for rendering. 
 We use a randomly placed orthographic camera, and a directional
light with a random direction within $60^\circ$ of the viewing direction to ensure
a sufficiently lit shape. All shapes are rendered with diffuse textureless surfaces,
along with self occlusion and shadows. In addition to the images, we also generate ground truth
surface normals.

\paragraph{Real images with ground truth} 
For both training and testing, we need a set of real-world images with ground truth of
surface normals.  For training, we need a validation set of real images to evaluate the
fitness of shapes, which is defined as how well they help the performance of a
shape-from-shading network on real images. For testing, we need a test set of real images
to evaluate the performance of the final network. 

We use the MIT-Berkeley Intrinsic Image dataset~\cite{BarronTPAMI2015,grosse2009ground}
as the source of real images. It includes images of 20 objects captured in a lab setting; 
each object has two images, one with texture and the other textureless. We use the
textureless version of the dataset because our method only evolves shape but not
texture. We adopt the official 50-50 train-test split, using the 10
training images as the validation set for fitness evaluation and the 10 test images to
evaluate the performance of the final network.

\paragraph{Setup of shape evolution}
In each iteration of shape evolution, the population size is maintained at $n=100$, and
$m=100$ new shapes are
composed. To select the shapes, 90\% of the population are sampled by a roulette wheel
where the probability of each shape being chosen is proportional to its fitness
score. The fitness score is the reciprocal of the mean angle error on the validation set.
The remaining 10\% are sampled using the diversity promoting strategy, where the shapes
are sampled also based on the rank $s$ of their computation graph size (from small to
large), with the relative selection probability set to $0.5^s$.

\paragraph{Evaluation Protocol}
To evaluate the shape-from-shading performance of the final network, we use standard
metrics proposed by prior work \cite{wang2015designing,BarronTPAMI2015}. 
We measure N-MAE and
N-MSE, \ie the mean angle distance (in radians) between the predicted normals and
ground-truth normals, and the mean squared errors of the normal vectors.  We also measure
the fraction of the pixels whose normals are within 11.25, 22.5, 30 degrees angle distance
of the ground-truth normals.

Since our network only
accepts 128$\times$128 input size but the images in the MIT-Berkeley dataset
have different sizes, we pad the images and scale them to 128$\times$128 to feed into the
network, and then scale them back and crop to the original sizes for evaluation.

\subsubsection{Baselines approaches}

We compare with a number of baseline approaches including ablated versions of our
algorithm. We describe them in detail below. 

\paragraph{SIRFS} SIRFS~\cite{BarronTPAMI2015} is an algorithm with state-of-the-art
performance on shape from shading. It is primarily based on optimization and manually
designed priors, with a small number of learned parameters. 
Our method only evolves shapes but not texture, so we
compare with SIRFS using the textureless images. Because the
published results~\cite{BarronTPAMI2015} only textured objects from the MIT-Berkeley
Intrinsic Image dataset, we obtained the results on textureless objects using 
 their open source code. 

\paragraph{Training with ShapeNet}
We also compare a baseline approach that trains the shape-from-shading network using
synthetic images rendered from an 
external shape dataset. We use a version of ShapeNet~\cite{shapenet2015v1}, a large dataset
of 3D CAD models that consists of approximately 51,300 shapes. We evaluate two variants of
this approach. 

\begin{itemize}
\setlength{\parskip}{0pt}
\setlength{\itemsep}{1pt} 
\item \emph{ShapeNet-vanilla} 
We train a single deep network on the synthetic images rendered using
shapes in ShapeNet.  Both the network structure and the rendering setting are the same as
in the evolutionary algorithm. For every $\tau$ RMSprop iterations (the number of iterations used
to fine-tune a network in the evolution algorithm), we record the validation
performance and save the snapshot of the network.  When testing, the snapshot with the
best validation performance is used. 

\item \emph{ShapeNet-incremental} Same as the first \emph{ShapeNet-incremental}, except
  that we restart the RMSprop training every $\tau$ iterations, initializing from the latest
  weights. This is because in our evolution algorithm only the network weights are
  reloaded for incremental training, while the RMSprop training starts from scratch. We
  include this baseline to eliminate any advantage the restarts might bring in our
  evolution algorithm.
\end{itemize}

\paragraph{Ablated versions of our algorithm} 
We consider three ablated versions of our algorithm: 
\begin{itemize}
\setlength{\parskip}{0pt}
\setlength{\itemsep}{1pt} 
\item \emph{Ours-no-feedback} The fitness score is replaced by a random value, while all other
parts of the algorithm remain unchanged.  The shapes are still being evolved, and the
networks are still being trained, but there is no feedback on how good the
shapes are.

\item \emph{Ours-no-evolution} The evolution is disabled, which means the population remains to
be the initial set of primitive shapes throughout the whole process.  This ablated
version is equivalent to training a set of networks on a fixed dataset and
picking the one from $n+m$ networks that has the best performance on the validation set
every $\tau$ training iterations. 

\item \emph{Ours-no-evolution-plus-ShapeNet} The evolution is disabled, and maintain a
  population of $n+m$ network instances being trained simultaneously.  For each $\tau$ iterations, the network with the best
validation performance is selected and copied to replace the entire population. It is equivalent
to \emph{Ours-no-evolution} except that the primitive
shapes are replaced by shapes randomly sampled from ShapeNet each time we render 
an image.  This ablation is to evaluate whether our evolved shapes are better than
ShapeNet shapes, controlling for any advantage our training algorithm might
have even without any evolution taking place.
\end{itemize}

\begin{table}[t]
\centering\setlength{\tabcolsep}{2pt}
\resizebox{\linewidth}{!}{
\begin{tabular}{@{}lccc|cc@{}}
\toprule
 & \multicolumn{3}{|c|}{Summary Stats $\uparrow$} & \multicolumn{2}{c}{Errors $\downarrow$} \\ \midrule
\multicolumn{1}{l|}{} & $\le11.25^\circ$ & $\le22.5^\circ$ & $\le30^\circ$ & MAE & MSE \\ \midrule
\multicolumn{1}{l|}{Random$^*$} & 1.9\% & 7.5\% & 13.1\% & 1.1627 & 1.3071 \\ \midrule
\multicolumn{1}{l|}{SIRFS~\cite{BarronTPAMI2015}} & 20.4\% & 53.3\% & 70.9\% & 0.4575 & 0.2964 \\ \midrule
\multicolumn{1}{l|}{ShapeNet-vanilla} & 12.7\% & 42.4\% & 62.8\% & 0.4831 & 0.2901 \\
\multicolumn{1}{l|}{ShapeNet-incremental} & 15.2\% & 48.4\% & 66.4\% & 0.4597 & 0.2717 \\ \midrule
\multicolumn{1}{l|}{Ours-no-evolution-plus-ShapeNet} & 14.2\% & 53.0\% & 72.1\% & 0.4232 & 0.2233 \\
\multicolumn{1}{l|}{Ours-no-evolution} & 17.3\% & 50.2\% & 66.1\% & 0.4673 & 0.2903 \\
\multicolumn{1}{l|}{Ours-no-feedback} & 19.1\% & 49.5\% & 66.3\% & 0.4477 & 0.2624 \\
\multicolumn{1}{l|}{Ours} & \textbf{21.6\%} & \textbf{55.5\%} & \textbf{73.5\%} & \textbf{0.4064} & \textbf{0.2204} \\ \bottomrule
\end{tabular}
}
\caption{The results of baselines and our approach on the test images.
$^*$Measured by uniformly randomly outputting unit vectors on the $+z$ hemisphere.
}
\label{tab:mbii_test}
\end{table}

\begin{figure*}[htb]
  \centering
  \includegraphics[width=\linewidth]{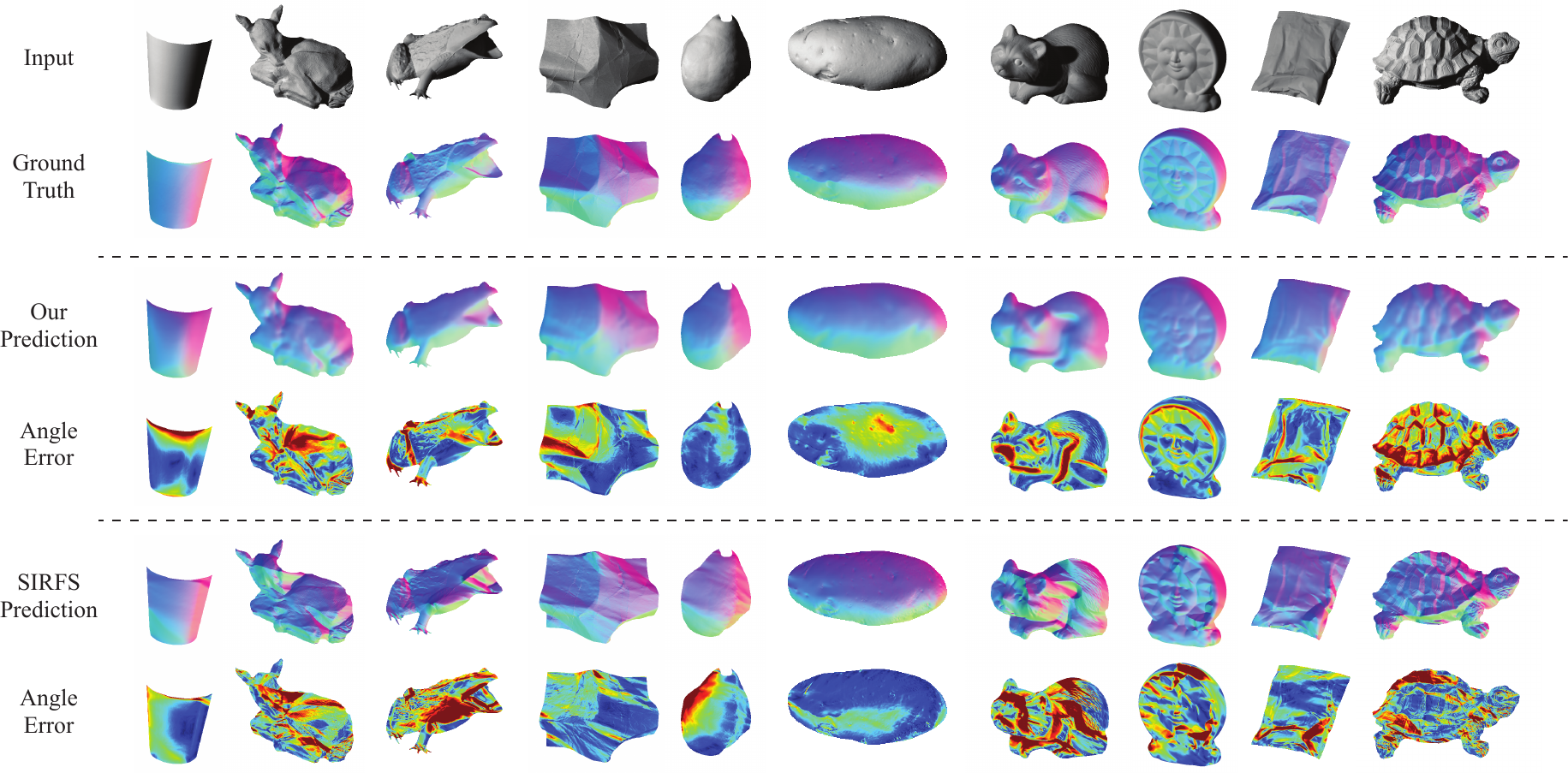}
  \caption{The qualitative results of our method and SIRFS~\cite{BarronTPAMI2015} on the test data.}
  \label{fig:mbii_visualization}
\end{figure*}
\subsubsection{Results and analysis}

\begin{figure}[htb]
    \centering
    \includegraphics[width=\columnwidth]{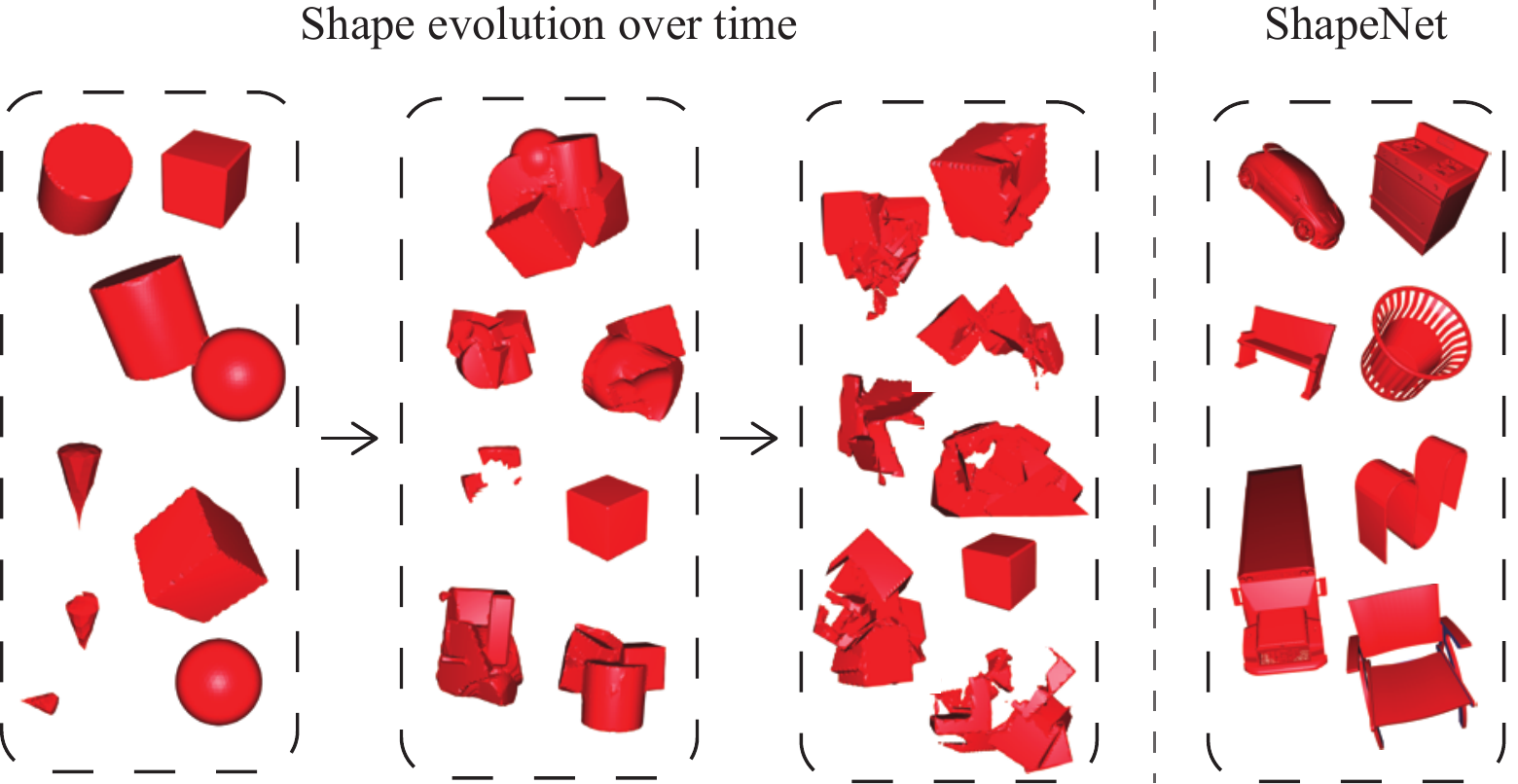}
    \caption{Example shapes at different stages of the evolution and shapes from ShapeNet.}
    \label{fig:mbii_population}
\end{figure}

Tab.~\ref{tab:mbii_test} compares the baselines with our approach. 
We first see that the deep network trained through shape evolution outperforms the
state-of-the-art SIRFS algorithm, without
using any external dataset except for the 10 training images in the MIT-Berkeley
dataset that are also used by SIRFS. 

We also see that our algorithm outperforms all baselines trained on ShapeNet as well as
all ablated versions. This shows that our approach can do
away with an external shape dataset and generate useful shapes from simple primitives,
and the evolved shapes are as useful as shapes from ShapeNet for this shape-from-shading
task. Fig.~\ref{fig:mbii_population} shows example shapes at different stages of the
evolution, as well as shapes from ShapeNet, and Fig.~\ref{fig:mbii_visualization} shows
qualitative results of our method and SIRFS on the test data.

More specifically, the \emph{Ours-no-evolution-plus-ShapeNet} ablation shows that our evolved shapes are
actually more useful than ShapeNet for the task, although this is not surprising given that the evolution is biased
toward being useful. Also it shows that the advantage our method has over using ShapeNet
is due to evolution, not idiosyncrasies of our training procedure. 

The \emph{Ours-no-evolution} ablation shows that our good performance is
not a result of well chosen primitive shapes, and evolution actually generates better
shapes. The \emph{Ours-no-feedback} ablation shows that the joint evolution and training is also
important---random evolution can produce complex shapes, but without guidance from
network training, the shapes are only slightly more useful than the primitives. 

\section{Conclusion}
We have introduced a new algorithm to jointly evolve 3D shapes and train a
shape-from-shading network through synthetic images.  We show that our approach can achieve
state-of-the-art performance on real images without using an external shape dataset.

\vspace{1em}
\noindent \textbf{Acknowledgments}
This work is partially supported by the National Science Foundation under Grant No.~1617767.

{\small
\bibliographystyle{ieee}

\begin{thebibliography}{10}\itemsep=-1pt

\bibitem{aubry2014seeing}
M.~Aubry, D.~Maturana, A.~Efros, B.~Russell, and J.~Sivic.
\newblock Seeing 3d chairs: exemplar part-based 2d-3d alignment using a large
  dataset of cad models.
\newblock In {\em CVPR}, 2014.

\bibitem{Bansal2016}
A.~Bansal, B.~Russell, and A.~Gupta.
\newblock Marr {R}evisited: 2{D}-3{D} model alignment via surface normal
  prediction.
\newblock In {\em CVPR}, 2016.

\bibitem{barron2011high}
J.~T. Barron and J.~Malik.
\newblock High-frequency shape and albedo from shading using natural image
  statistics.
\newblock In {\em Computer Vision and Pattern Recognition (CVPR), 2011 IEEE
  Conference on}, pages 2521--2528. IEEE, 2011.

\bibitem{barron2012shape}
J.~T. Barron and J.~Malik.
\newblock Shape, albedo, and illumination from a single image of an unknown
  object.
\newblock In {\em Computer Vision and Pattern Recognition (CVPR), 2012 IEEE
  Conference on}, pages 334--341. IEEE, 2012.

\bibitem{BarronTPAMI2015}
J.~T. Barron and J.~Malik.
\newblock Shape, illumination, and reflectance from shading.
\newblock {\em TPAMI}, 2015.

\bibitem{butler2012a}
D.~J. Butler, J.~Wulff, G.~B. Stanley, and M.~J. Black.
\newblock A naturalistic open source movie for optical flow evaluation.
\newblock In {A. Fitzgibbon et al. (Eds.)}, editor, {\em European Conf. on
  Computer Vision (ECCV)}, Part IV, LNCS 7577, pages 611--625. Springer-Verlag,
  Oct. 2012.

\bibitem{chakrabarti2016depth}
A.~Chakrabarti, J.~Shao, and G.~Shakhnarovich.
\newblock Depth from a single image by harmonizing overcomplete local network
  predictions.
\newblock In D.~D. Lee, M.~Sugiyama, U.~V. Luxburg, I.~Guyon, and R.~Garnett,
  editors, {\em Advances in Neural Information Processing Systems 29}, pages
  2658--2666. Curran Associates, Inc., 2016.

\bibitem{shapenet2015v1}
A.~X. Chang, T.~Funkhouser, L.~Guibas, P.~Hanrahan, Q.~Huang, Z.~Li,
  S.~Savarese, M.~Savva, S.~Song, H.~Su, J.~Xiao, L.~Yi, and F.~Yu.
\newblock {ShapeNet: An Information-Rich 3D Model Repository}.
\newblock Technical Report arXiv:1512.03012 [cs.GR], Stanford University ---
  Princeton University --- Toyota Technological Institute at Chicago, 2015.

\bibitem{choi2016a}
S.~Choi, Q.-Y. Zhou, S.~Miller, and V.~Koltun.
\newblock A large dataset of object scans.
\newblock {\em arXiv:1602.02481}, 2016.

\bibitem{choy20163d}
C.~B. Choy, D.~Xu, J.~Gwak, K.~Chen, and S.~Savarese.
\newblock 3d-r2n2: A unified approach for single and multi-view 3d object
  reconstruction.
\newblock In {\em Proceedings of the European Conference on Computer Vision
  ({ECCV})}, 2016.

\bibitem{clune2011evolving}
J.~Clune and H.~Lipson.
\newblock Evolving 3d objects with a generative encoding inspired by
  developmental biology.
\newblock {\em SIGEVOlution}, 5(4):2--12, Nov. 2011.

\bibitem{cole2012shapecollage}
F.~Cole, P.~Isola, W.~T. Freeman, F.~Durand, and E.~H. Adelson.
\newblock Shapecollage: Occlusion-aware, example-based shape interpretation.
\newblock In {\em Computer Vision--ECCV 2012}, pages 665--678. Springer, 2012.

\bibitem{criminisi2000shape}
A.~Criminisi and A.~Zisserman.
\newblock Shape from texture: Homogeneity revisited.
\newblock In {\em BMVC}, pages 1--10, 2000.

\bibitem{dai2017scannet}
A.~Dai, A.~X. Chang, M.~Savva, M.~Halber, T.~Funkhouser, and M.~Nie{\ss}ner.
\newblock Scannet: Richly-annotated 3d reconstructions of indoor scenes.
\newblock In {\em Proc. Computer Vision and Pattern Recognition (CVPR), IEEE},
  2017.

\bibitem{ecker2010polynomial}
A.~Ecker and A.~D. Jepson.
\newblock Polynomial shape from shading.
\newblock In {\em 2010 IEEE Computer Society Conference on Computer Vision and
  Pattern Recognition}, pages 145--152, June 2010.

\bibitem{Eigen2015}
D.~Eigen and R.~Fergus.
\newblock {Predicting depth, surface normals and semantic labels with a common
  multi-scale convolutional architecture}.
\newblock In {\em ICCV}, 2015.

\bibitem{Geiger2012CVPR}
A.~Geiger, P.~Lenz, and R.~Urtasun.
\newblock Are we ready for autonomous driving? the kitti vision benchmark
  suite.
\newblock In {\em Conference on Computer Vision and Pattern Recognition
  (CVPR)}, 2012.

\bibitem{grosse2009ground}
R.~Grosse, M.~K. Johnson, E.~H. Adelson, and W.~T. Freeman.
\newblock Ground truth dataset and baseline evaluations for intrinsic image
  algorithms.
\newblock In {\em Computer Vision, 2009 IEEE 12th International Conference on},
  pages 2335--2342. IEEE, 2009.

\bibitem{hoiem2007recovering}
D.~Hoiem, A.~N. Stein, A.~Efros, M.~Hebert, et~al.
\newblock Recovering occlusion boundaries from a single image.
\newblock In {\em Computer Vision, 2007. ICCV 2007. IEEE 11th International
  Conference on}, pages 1--8. IEEE, 2007.

\bibitem{holland1975adaptation}
J.~H. Holland.
\newblock {\em Adaptation in Natural and Artificial Systems}.
\newblock University of Michigan Press, Ann Arbor, MI, 1975.
\newblock second edition, 1992.

\bibitem{Mitsuba}
W.~Jakob.
\newblock Mitsuba renderer, 2010.
\newblock http://www.mitsuba-renderer.org.

\bibitem{janoch2013category}
A.~Janoch, S.~Karayev, Y.~Jia, J.~T. Barron, M.~Fritz, K.~Saenko, and
  T.~Darrell.
\newblock A category-level 3d object dataset: Putting the kinect to work.
\newblock In {\em Consumer Depth Cameras for Computer Vision}, pages 141--165.
  Springer, 2013.

\bibitem{lorensen1987marching}
W.~E. Lorensen and H.~E. Cline.
\newblock Marching cubes: A high resolution 3d surface construction algorithm.
\newblock In {\em Proceedings of the 14th Annual Conference on Computer
  Graphics and Interactive Techniques}, SIGGRAPH '87, pages 163--169, New York,
  NY, USA, 1987. ACM.

\bibitem{massa2016deep}
F.~Massa, B.~Russell, and M.~Aubry.
\newblock Deep exemplar 2d-3d detection by adapting from real to rendered
  views.
\newblock In {\em Conference on Computer Vision and Pattern Recognition
  (CVPR)}, 2016.

\bibitem{mccormac2017scenenet}
J.~McCormac, A.~Handa, S.~Leutenegger, and A.~J. Davison.
\newblock Scenenet rgb-d: Can 5m synthetic images beat generic imagenet
  pre-training on indoor segmentation?
\newblock In {\em The IEEE International Conference on Computer Vision (ICCV)},
  Oct 2017.

\bibitem{Silberman:ECCV12}
P.~K. Nathan~Silberman, Derek~Hoiem and R.~Fergus.
\newblock Indoor segmentation and support inference from rgbd images.
\newblock In {\em ECCV}, 2012.

\bibitem{newell2016stacked}
A.~Newell, K.~Yang, and J.~Deng.
\newblock Stacked hourglass networks for human pose estimation.
\newblock In B.~Leibe, J.~Matas, N.~Sebe, and M.~Welling, editors, {\em
  Computer Vision - {ECCV} 2016 - 14th European Conference, Amsterdam, The
  Netherlands, October 11-14, 2016, Proceedings, Part {VIII}}, volume 9912 of
  {\em Lecture Notes in Computer Science}, pages 483--499. Springer, 2016.

\bibitem{ricci1973constructive}
A.~Ricci.
\newblock A constructive geometry for computer graphics.
\newblock {\em The Computer Journal}, 16(2):157--160, 1973.

\bibitem{richter2017playing}
S.~R. Richter, Z.~Hayder, and V.~Koltun.
\newblock Playing for benchmarks.
\newblock In {\em The IEEE International Conference on Computer Vision (ICCV)},
  Oct 2017.

\bibitem{richter2015discriminative}
S.~R. Richter and S.~Roth.
\newblock Discriminative shape from shading in uncalibrated illumination.
\newblock In {\em 2015 IEEE Conference on Computer Vision and Pattern
  Recognition (CVPR)}, pages 1128--1136, June 2015.

\bibitem{richter2016playing}
S.~R. Richter, V.~Vineet, S.~Roth, and V.~Koltun.
\newblock Playing for data: {G}round truth from computer games.
\newblock In B.~Leibe, J.~Matas, N.~Sebe, and M.~Welling, editors, {\em
  European Conference on Computer Vision (ECCV)}, volume 9906 of {\em LNCS},
  pages 102--118. Springer International Publishing, 2016.

\bibitem{saxena2009make3d}
A.~Saxena, M.~Sun, and A.~Y. Ng.
\newblock Make3d: Learning 3d scene structure from a single still image.
\newblock {\em Pattern Analysis and Machine Intelligence, IEEE Transactions
  on}, 31(5):824--840, 2009.

\bibitem{song2015sun}
S.~Song, S.~P. Lichtenberg, and J.~Xiao.
\newblock Sun rgb-d: A rgb-d scene understanding benchmark suite.
\newblock In {\em Proceedings of the IEEE Conference on Computer Vision and
  Pattern Recognition}, pages 567--576, 2015.

\bibitem{stanley2007compositional}
K.~O. Stanley.
\newblock Compositional pattern producing networks: A novel abstraction of
  development.
\newblock {\em Genetic programming and evolvable machines}, 8(2):131--162,
  2007.

\bibitem{stanley2002evolving}
K.~O. Stanley and R.~Miikkulainen.
\newblock Evolving neural networks through augmenting topologies.
\newblock {\em Evolutionary Computation}, 10(2):99--127, 2002.

\bibitem{su2015render}
H.~Su, C.~R. Qi, Y.~Li, and L.~J. Guibas.
\newblock Render for cnn: Viewpoint estimation in images using cnns trained
  with rendered 3d model views.
\newblock In {\em Proceedings of the IEEE International Conference on Computer
  Vision}, pages 2686--2694, 2015.

\bibitem{tatarchenko2016multi}
M.~Tatarchenko, A.~Dosovitskiy, and T.~Brox.
\newblock Multi-view 3d models from single images with a convolutional network.
\newblock In {\em European Conference on Computer Vision (ECCV)}, 2016.

\bibitem{tieleman2012lecture}
T.~Tieleman and G.~Hinton.
\newblock Lecture 6.5-rmsprop: Divide the gradient by a running average of its
  recent magnitude.
\newblock {\em COURSERA: Neural networks for machine learning}, 4(2):26--31,
  2012.

\bibitem{wang2015towards}
P.~Wang, X.~Shen, Z.~Lin, S.~Cohen, B.~Price, and A.~L. Yuille.
\newblock Towards unified depth and semantic prediction from a single image.
\newblock In {\em Proceedings of the IEEE Conference on Computer Vision and
  Pattern Recognition}, pages 2800--2809, 2015.

\bibitem{wang2015designing}
X.~Wang, D.~Fouhey, and A.~Gupta.
\newblock Designing deep networks for surface normal estimation.
\newblock In {\em Proceedings of the IEEE Conference on Computer Vision and
  Pattern Recognition}, pages 539--547, 2015.

\bibitem{wu20153d}
Z.~Wu, S.~Song, A.~Khosla, F.~Yu, L.~Zhang, X.~Tang, and J.~Xiao.
\newblock 3d shapenets: A deep representation for volumetric shapes.
\newblock In {\em Proceedings of the IEEE Conference on Computer Vision and
  Pattern Recognition}, pages 1912--1920, 2015.

\bibitem{xiang2016objectnet3d}
Y.~Xiang, W.~Kim, W.~Chen, J.~Ji, C.~Choy, H.~Su, R.~Mottaghi, L.~Guibas, and
  S.~Savarese.
\newblock Objectnet3d: A large scale database for 3d object recognition.
\newblock In {\em European Conference Computer Vision (ECCV)}, 2016.

\bibitem{xiong2015shading}
Y.~Xiong, A.~Chakrabarti, R.~Basri, S.~J. Gortler, D.~W. Jacobs, and
  T.~Zickler.
\newblock From shading to local shape.
\newblock {\em Pattern Analysis and Machine Intelligence, IEEE Transactions
  on}, 37(1):67--79, 2015.

\bibitem{zhang1999shape}
R.~Zhang, P.-S. Tsai, J.~E. Cryer, and M.~Shah.
\newblock Shape from shading: A survey.
\newblock {\em IEEE Trans. Pattern Anal. Mach. Intell.}, 21(8):690--706, Aug.
  1999.

\bibitem{zhang2016physically}
Y.~Zhang, S.~Song, E.~Yumer, M.~Savva, J.-Y. Lee, H.~Jin, and T.~Funkhouser.
\newblock Physically-based rendering for indoor scene understanding using
  convolutional neural networks.
\newblock {\em The IEEE Conference on Computer Vision and Pattern Recognition
  (CVPR)}, 2017.

\end{thebibliography}

}

\end{document}